\documentclass[letterpaper]{article} % DO NOT CHANGE THIS
\usepackage[preprint]{aaai2027}  % DO NOT CHANGE THIS
\usepackage[hyphens]{url}  % DO NOT CHANGE THIS
\usepackage{graphicx} % DO NOT CHANGE THIS
\usepackage{natbib}  % DO NOT CHANGE THIS AND DO NOT ADD ANY OPTIONS TO IT
\usepackage{caption} % DO NOT CHANGE THIS AND DO NOT ADD ANY OPTIONS TO IT
\usepackage{algorithm}
\usepackage{algorithmic}

\usepackage{newfloat}
\usepackage{listings}
\DeclareCaptionStyle{ruled}{labelfont=normalfont,labelsep=colon,strut=off} % DO NOT CHANGE THIS
\floatstyle{ruled}
\newfloat{listing}{tb}{lst}{}
\floatname{listing}{Listing}

\usepackage{tabularx, booktabs}
\newcommand{\tna}{\textendash}  % table N/A (em dash)

\newcommand{\dataname}[1]{\textsc{#1}}
\newcommand{\best}[1]{\textbf{#1}}              % legacy alias
\newcommand{\bestsup}[1]{\textbf{#1}}           % best supervised (trained head / finetune)
\newcommand{\besttf}[1]{\underline{\textbf{#1}}} % best training-free (frozen weights)
\newcommand{\ind}[1]{\operatorname{\mathbb{I}}\!\left(#1\right)} % indicator: 1 if #1, else 0

\usepackage{amsmath,amssymb}
\usepackage{graphicx}
\usepackage{booktabs}
\usepackage{longtable}
\usepackage{multirow}
\usepackage{caption}
\usepackage{subcaption}
\usepackage{xcolor}

\usepackage[expansion=false]{microtype}
\usepackage{enumitem}
\setlist{nosep,leftmargin=1.4em}
\title{Sparse Concept Channels in Frozen 3D CT Vision Encoders}
\author {
    Farhad Nooralahzadeh\corresponding\textsuperscript{\rm 1,\rm 2},
    Lea Bogensperger\textsuperscript{\rm 1},
     Christian Bluethgen\textsuperscript{\rm 3},
     Michael Krauthammer\textsuperscript{\rm 1} 
}
\affiliations {
    \textsuperscript{\rm 1}Department of Quantitative Biomedicine, University of Zurich, Switzerland\\
    \textsuperscript{\rm 2} Institute of Computer Science, Zurich University of Applied Sciences, Switzerland\\
    \textsuperscript{\rm 3}Center for Artificial Intelligence in Medicine and Imaging, Stanford University, USA\\
    farhad.nooralahzadeh@uzh.ch
}
\begin{document}

\maketitle

\begin{abstract}
Large vision-language models are becoming increasingly dominant in 3D medical image interpretation, but we rarely know \textit{which} internal units encode clinical findings or \textit{where} that information lives in the representation. We first study this on a 3D chest vision-language model (\dataname{Pillar-0}) by probing its frozen vision embeddings. We show that (i) each radiological finding is encoded by a \emph{sparse} set of $\sim$10 vision-encoder channels that match full-feature classification performance and far exceed a zero-shot text prompting; (ii) turning off the channels tied to one finding, that finding’s score collapses while unrelated labels stay stable; and (iii) the same sparse probe \emph{replicates} on an architecturally unrelated 3D abdominal VLM (\dataname{Merlin}) suggesting a general property of frozen medical encoders. Our training-free concept channel probe (CCP) method, paired with a corpus-derived report template, outperforms published CT-CHAT on clinical efficacy and NLG metrics (F1 0.549 vs. 0.184; BLEU 0.483 vs. 0.373) at 22$\times$lower latency. Our results provide a clear, reproducible characterization of how frozen medical encoders represent findings, demonstrating direct applicability across models. %that the same CCP-to-report pipeline can be applied directly and effectively across different models.
\end{abstract}


\section{Introduction}
Foundation models for medical imaging are now strong enough to be considered as clinical decision support tools, yet their internal representations are poorly understood. We explore these representations by following the questions that matter in terms of explainability:
\emph{What} concepts does a vision encoder encode, and \emph{where}, in which units and at what
spatial resolution, does that information reside? Answering them without fine-tuning is attractive
because it (a) avoids the cost and risk of retraining a large model, and (b) yields explanations
about the \emph{deployed} model rather than a modified copy.
\begin{figure*}[t]
    \centering
    \includegraphics[width=\linewidth]{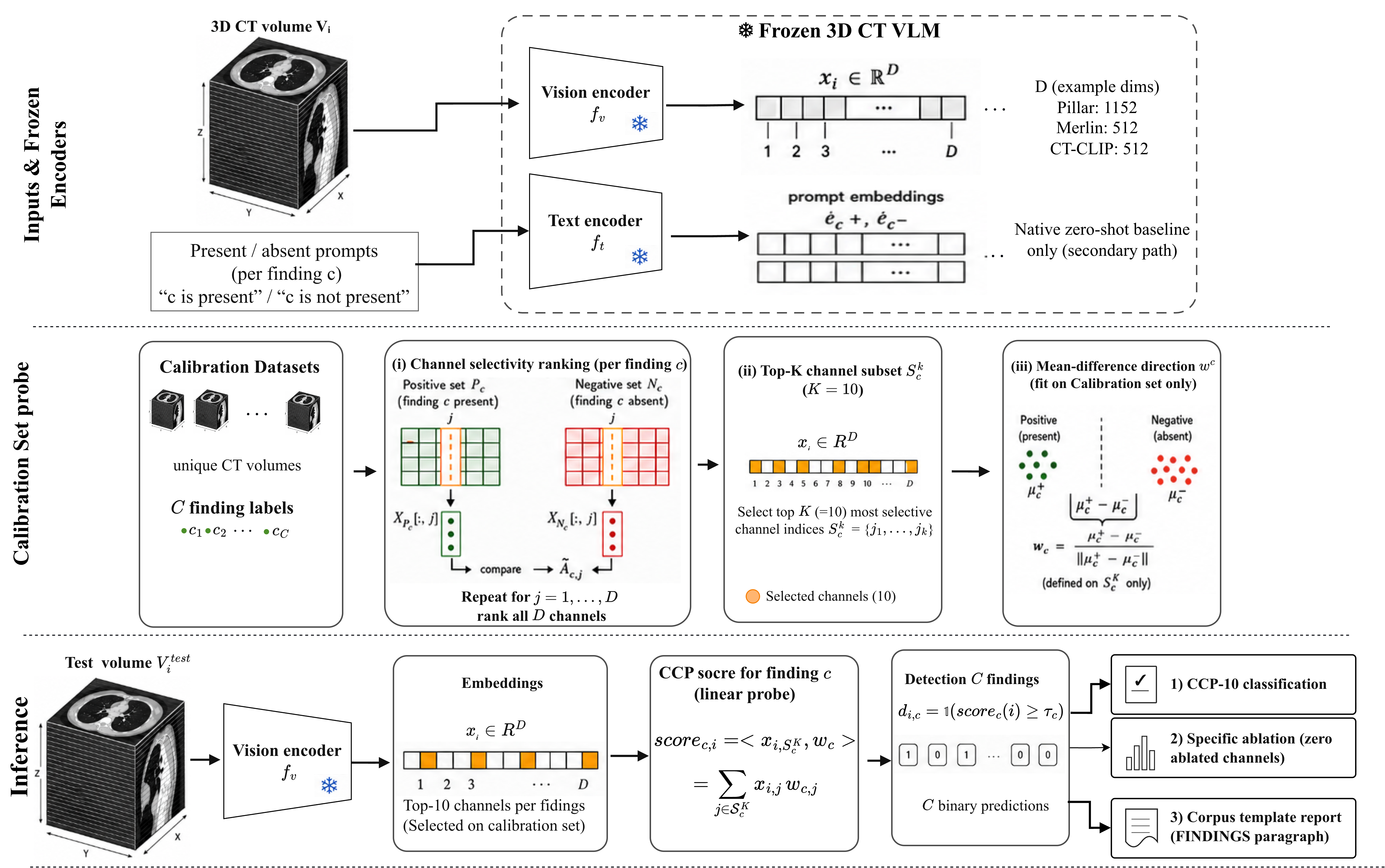}
    \caption{Concept Channel Probe (CCP) overview.
    % A frozen 3D vision-language model encodes CT volumes to embeddings $x_i \in \mathbb{R}^{D}$. CCP uses multi-label annotations as statistical probes to rank channels by selectivity, extract sparse top-$K$ detectors per finding, and calibrate thresholds. The resulting sparse channels enable classification, causal ablation, and deterministic report generation—all without training.
    }
    \label{fig:pipeline}
\end{figure*}
We therefore take a deliberately \emph{training-free} approach. As shown in Figure \ref{fig:pipeline}, we freeze a 3D vision-language model and cache its final image embedding $x_i$. We then use multi-label volume annotations as statistical \emph{probes} on $x_i$ - not to train a network, but to ask: which 
embedding dimensions linearly encode clinical findings? % where concept information is linearly available. 
% We design the concept channel probe (CCP) to rank channels by development set separability, form a mean-difference detector on the top-K coordinates, and threshold on a held-out development split. In our design there is no backpropagation and no change to deployed weights.
This leads to three contributions:
\textbf{ (1) Sparsity} Each radiological finding is encoded by $\sim$10 channels 
that achieve comparable classification performance to the full D-dimensional embedding, %For each CT finding $\sim$10 channels of the frozen feature classify the finding as well as the full-D dimensional ($D\gg10$) feature.
\textbf{(2) Causality} Zeroing a finding's specific channels reduces its own probe AUROC score $\sim$20$\times$ more than other findings, and the specific channels cluster into clinically coherent groups; and \textbf{ (3) Generalization across backbones} The sparse organization replicates on another 3D CT backbone (3D VLM with a different pretraining pipeline), indicating the phenomenon is a property of frozen VLMs rather than of one architecture.

We then show the probe is \emph{useful} and \emph{backbone-portable}: the same CCP-10 detections and corpus-derived template verbalizer outperform trained CT VLMs on clinical and
factual metrics at a fraction of the cost on underlying 3D VLMs.

\section{Related Work}

\textbf{3D medical vision-language models.}
Recent foundation models pair 3D CT encoders with text supervision, 
enabling zero-shot classification and report generation. Examples 
include CT-CHAT~\cite{hamamci2026generalist} and Pillar-0~\cite{agrawal2025pillar} for chest CT, and Merlin~\cite{blankemeier2024merlin} for abdominal 
CT. These approaches optimize \emph{what} the model predicts through 
pretraining and decoding. In contrast, we treat frozen encoders as 
interpretable systems and ask: \emph{where} does finding information reside 
in the embeddings? We answer this through statistical probes rather 
than weight updates.

\noindent\textbf{Probing and interpretability of learned representations.} Linear probes~\cite{alain2016understanding} test what information is linearly decodable from frozen features, while network dissection~\cite{bau2017network} aligns individual units with human concepts. Superposition analyses~\cite{elhage2022superposition} show that features can be distributed across many coordinates, and representation-engineering methods~\cite{zou2023representation} read and steer concepts via mean-difference directions. Our concept channel probe (CCP) adapts these ideas to 3D medical encoders: we rank channels by calibration-set selectivity and fit a closed-form mean-difference detector on a sparse top-$K$ subset, yielding a per-finding, training-free probe.

\noindent\textbf{Causal localization.}
Causal mediation analysis~\cite{NEURIPS2019_2c601ad9} and activation-level interventions for locating and editing knowledge in networks~\cite{meng2022locating} establish necessity by ablating internal components. We apply causal intervention to frozen embeddings: zeroing each finding's top-$K$ channels selectively impacts its score, leaving others unchanged.

\noindent\textbf{Radiology report generation and evaluation.}
Report evaluation uses clinical metrics (RadGraph-F1~\cite{delbrouck-etal-2024-radgraph}, 
RadEval~\cite{xu-etal-2025-radeval}).
%Report-generation systems are increasingly scored with clinically grounded metrics such as
%RadGraph-F1~\cite{delbrouck-etal-2024-radgraph} and unified suites like
%RadEval~\cite{xu-etal-2025-radeval}. 
Rather than end-to-end generation, we decouple detection from
verbalization: CCP produces reliable binary detections that a deterministic corpus template turns into
FINDINGS text, which we evaluate under the same clinical and surface metrics.

\section{Method}\label{sec:method}
\label{sec:methodology}
We avoid weight fine-tuning and opaque generative decoding by 
taking a training-free approach: each 3D image passes through a \emph{frozen} vision encoder to produce 
embedding $x_i$, on which all probes (CCP-$K$, sparsity, causal ablation, 
reports) operate without training. 

\subsection{Frozen Backbone Inputs}
\label{sec:backbone_inputs}
Let $V_i$ denote a CT volume. Each backbone applies its own preprocessing; 
we cache the resulting frozen embedding per scan:
\begin{equation}
x_i = f_V\!\big(\mathrm{preprocess}_V(V_i)\big) \in \mathbb{R}^{D}
\label{eq:backbone_input}
\end{equation}
where $f_V$ and $D$ are fixed by the pretrained model (e.g., \dataname{Pillar-0}). 
We treat $x_i$ as a black-box input; backbone internals remain frozen. Our 
channel-ranking and ablation logic generalizes across architectures; only 
$D$ and preprocessing differ.

\subsection{Concept Channel Probe (CCP-$K$)}
\label{sec:ccp}

\textbf{CCP-$K$} (\emph{concept channel probe}) is our primary linear probe on frozen channels. Given a multi-label task with $C$ findings, for each finding $c\in\{1,\ldots,C\}$, we (i)~rank channel coordinates $j$ by selectivity score $\tilde{A}_{c,j}$ on calibration set
(Eq.~\ref{eq:sel_formal}) and form top-$K$ sets $\mathcal{S}^K_c$;
(ii)~fit a closed-form mean-difference direction $w_c$ on each $\mathcal{S}^K_c$ (Eq.~\ref{eq:topk_formal});
(iii)~compute continuous scores $\text{score}_c(i)$, calibrate thresholds $\tau_c$ on a held-out \emph{calibration} split, and design binary detections $d_i$ (Eq.~\ref{eq:detections}); and
(iv) further analyze finding-specific channels and causal ablation
(Eqs.~\ref{eq:abl_formal}). The calibration split is used for channel ranking ($w_c$), score calibration, and $\tau_c$.
Here the finding index is $c\in\{1,\ldots,C\}$; channel index is $j\in\{1,\ldots,D\}$; circuit size $K$ is a hyperparameter fixed before evaluation.
$\ind{\cdot}$ is the indicator function ($1$ if the condition holds, $0$ otherwise).

\paragraph{Channel ranking and scoring.}
For pathology $c$, let $P_c=\{i:\text{finding }c\text{ present}\}$ and
$N_c=\{i:\text{finding }c\text{ absent}\}$ on the calibration set.
We define per-coordinate selectivity following network dissection~\cite{bau2017network}:
\begin{equation}
A_{c,j}=\frac{1}{|P_c||N_c|}\!\!\sum_{p\in P_c}\sum_{n\in N_c}\!
\ind{x_{p,j}>x_{n,j}},\quad
\tilde A_{c,j}=A_{c,j}-\tfrac12
\label{eq:sel_formal}
\end{equation}
$A_{c,j}$ is the pairwise win rate (equivalent to per-channel AUROC) and is invariant to finding prevalence—it measures separability, not class ratio. Centering to $\tilde A_{c,j}$ puts chance at zero. This makes ranking robust to the severe class imbalance in multi-label CT (prevalence ranges $\sim$7\% to $\sim$45\%).

We rank channels by $\tilde A_{c,\cdot}$ and extract the top-$K$ set $\mathcal{S}^K_c$, then fit a mean-difference projection following~\cite{zou2023representation} to score unseen volumes as a sparse alternative to fine-tuning~\cite{hamamci2026generalist}:
\begin{equation}
\begin{aligned}
\small
\underbrace{w_c=\frac{\mu^{P}_{\mathcal{S}^K_c}-\mu^{N}_{\mathcal{S}^K_c}}
{\lVert \mu^{P}_{\mathcal{S}^K_c}-\mu^{N}_{\mathcal{S}^K_c}\rVert}}_{\text{fit on calibration set}},\quad
\underbrace{\text{score}_c(i)=\big\langle x_{i,\mathcal{S}^K_c},\,w_c\big\rangle
=\!\!\sum_{j\in\mathcal{S}^K_c}\!\! x_{i,j}\, w_{c,j}}_{\text{applied per volume}}
\end{aligned}
\label{eq:topk_formal}
\end{equation}
Here $\mu^{P}_{\mathcal{S}^K_c}$ and $\mu^{N}_{\mathcal{S}^K_c}$ are class means over calibration positives and negatives, restricted to top-$K$ channels. This mean-difference approach avoids bias from class imbalance, unlike pooled or least-squares fits.
Critically, $\mathcal{S}^K_c$ and $w_c$ are computed only on the calibration set. At inference, both are frozen; scoring requires extracting $x_{i,\mathcal{S}^K_c}$ and computing its dot product with $w_c$. The test volume never enters $\mu^{P}$ or $\mu^{N}$, ensuring no data leakage.

\paragraph{Binary detections.}
We turn each continuous probe score into a present/absent prediction by comparing
$\text{score}_c(i)$ to a per-finding cutoff $\tau_c$:
\begin{equation}
d_{i,c}=\ind{\text{score}_c(i)\ge\tau_c},\quad
d_i=(d_{i,1},\ldots,d_{i,C})\in\{0,1\}^{C}
\label{eq:detections}
\end{equation}
where $d_{i,c}{=}1$ means CCP predicts finding $c$ on volume $i$; $d_i$ is also the detection vector fed to the report generation process afterwards.
% On the calibration split we pick each $\tau_c$ by sweeping candidate cutoffs on calibration score and
% choose the value that maximizes per-finding F1; we then freeze
% $\{\mathcal{S}^K_c,w_c,\tau_c\}_{c=1}^{C}$ and apply the same rule on test split without refitting.
On the calibration split we select each $\tau_c$ by sweeping candidate cutoffs on the calibration scores. We use two operating points depending on the downstream task: for the report-generation detections $d_i$ we choose the cutoff that maximizes per-finding F1, whereas for the classification benchmark (Table~\ref{tab:ctrate_bench}) we follow the \dataname{CT-CLIP} protocol and place $\tau_c$ at the dev ROC point closest to the top-left corner---a prevalence-independent operating point that does not depend on the positive/negative ratio. In both cases we then freeze $\{\mathcal{S}^K_c,w_c,\tau_c\}_{c=1}^{C}$ and apply the same rule on the test split without refitting.

\paragraph{Causal ablation.}
The CCP-$K$ classification selects the global top-$K$ coordinates for each finding. 
To investigate whether these channels are \emph{necessary} (i.e., causally responsible for diagnostic scores), we perform controlled ablation following~\citet{NEURIPS2019_2c601ad9} and ~\citet{meng2022locating}: for source finding $c$, define $\mathcal{S}^{\mathrm{abl}}_c$ as the top-$K$ coordinates with largest $\tilde A_{c,j}$ on the calibration split.
We set $x'_{i,j}{=}0$ for $j\in\mathcal{S}^{\mathrm{abl}}_c$ before recomputing frozen probe scores on a held-out split. The drop matrix is: 
\begin{equation}
\begin{aligned}
\Delta_{c\to c'}&=
\mathrm{AUROC}_{c'}\!\big(\{x_i\}\big)-\mathrm{AUROC}_{c'}\!\big(\{x'_i\}\big),\\
\text{sel}&=\frac{1}{C}\sum_{c=1}^{C}\left(
\Delta_{c\to c}-\frac{1}{C-1}\sum_{c'\ne c}\Delta_{c\to c'}\right)
\end{aligned}
\label{eq:abl_formal}
\end{equation}
where $C$ is the number of findings.
%For ablation source finding $c$, let $\mathcal{S}_c$ be the top-$K$ finding-specific channels, fit on the calibration split.
For each source $c$, the parenthesis in Eq.~\ref{eq:abl_formal} is the gap between the \emph{on-target} drop
$\Delta_{c\to c}$ and the mean \emph{off-target} drop over $c'\ne c$, and \texttt{sel} averages that gap over all $c$.
Large \texttt{sel} means a finding's specific channels are \emph{necessary} for its own score but largely \emph{unnecessary} for others.

\subsection{Training-free report generation}
\label{sec:report_method}
We generate reports deterministically from $d_i$ (Eq.~\ref{eq:detections}) 
using a corpus-derived template. From training reports (excluding calibration 
examples), we extract: (i)~$T_0$, the most frequent FINDINGS report (a common 
"normal chest" skeleton); and (ii)~$\{u_c\}$, the most common positive sentence 
per finding $c$ with matching keywords. 
At test time, we edit $T_0$: (1)~split sentences, tag keywords, mark 
negation; (2)~remove sentences for absent findings, append $u_c$ for 
$d_{i,c}=1$, generate IMPRESSION.
The verbalizer never introduces findings beyond $d_i$ (no hallucinated positives).

\begin{table}[t]
\centering
\small
\setlength{\tabcolsep}{1pt}
\begin{tabular}{@{}l|l|cccc@{}}
\toprule
Dataset & method & AUROC & F1 & Accuracy & PR \\
\midrule
\multirow{10}{*}{\shortstack[l]{\scriptsize\dataname{CT-RATE} \\($n{=}1564$)}}
  & Random & 0.500 & 0.566 & 0.507 & 0.189 \\
  & CT-CLIP zero-shot$^\S$ & 0.731 & 0.707 & 0.668 & 0.323 \\
  & CT-CLIP ClassFine$^\S$ & 0.756 & 0.724 & 0.689 & 0.339 \\
  & CT-CLIP VocabFine$^\S$ & 0.756 & 0.738 & 0.705 & 0.353 \\
  & \dataname{Pillar-0} ClassFine$^\dagger$ & \bestsup{0.833} & \bestsup{0.800} & \bestsup{0.778} & \bestsup{0.431} \\
  & \dataname{Pillar-0} zero-shot$^\ast$ & 0.682 & 0.693 & 0.652 & 0.291 \\
  % &  \dataname{Merlin} zero-shot $^\ast$ & 0.571 & 0.598 & 0.545 & 0.218 \\
  & CCP-Full--- \dataname{Pillar-0} & 0.793 & 0.782 & 0.760 & 0.400 \\
  & CCP-10 --- \dataname{Pillar-0}$^\ddagger$ & \besttf{0.798} & \besttf{0.790} & \besttf{0.768} & \besttf{0.405} \\
  % & CCP-10 --- \dataname{Merlin}$^\ddagger$ & 0.729 & 0.734 & 0.703 & 0.337 \\
\midrule
\multirow{10}{*}{\shortstack[l]{\scriptsize\dataname{RadChest-CT}\\($n{=}3630$)}}
  & Random & 0.505 & 0.565 & 0.506 & 0.245 \\
  & CT-CLIP zero-shot$^\S$ & 0.629 & 0.637 & 0.592 & 0.335 \\
  & CT-CLIP ClassFine$^\S$ & 0.643 & 0.644 & 0.599 & 0.346 \\
  & CT-CLIP VocabFine$^\S$ & 0.650 & \bestsup{0.677} & \bestsup{0.636} & \bestsup{0.346} \\
  & \dataname{Pillar-0} ClassFine$^\dagger$ & \bestsup{0.713} & 0.609 & 0.570 & 0.335 \\
  & \dataname{Pillar-0} zero-shot $^\ast$ & 0.634 & 0.579 & 0.533 & 0.295 \\
  % & \dataname{Merlin} zero-shot$^\ast$ & 0.504 & 0.394 & 0.381 & 0.240 \\
  & CCP-Full --- \dataname{Pillar-0} & 0.683 & 0.596 & 0.553 & 0.324 \\
  & CCP-10 --- \dataname{Pillar-0}$^\ddagger$ & \besttf{0.686} & \besttf{0.622} & \besttf{0.581} & \besttf{0.333} \\
  % & CCP-10 --- \dataname{Merlin}$^\ddagger$ & 0.602 & 0.248 & 0.287 & 0.251 \\
\bottomrule
\end{tabular}
\caption{Multi-abnormality classification (classification metrics: mean AUROC;
mean per-finding weighted F1, accuracy, and precision (PR) after ROC upper-left thresholds on calibration set: \dataname{CT-RATE} official test set ($n{=}1564$) and \dataname{RadChest-CT} ($n{=}3630$).
$^\S$\dataname{CT-CLIP} rows \textbf{extracted} from ~\citet{hamamci2026generalist};
$^\dagger$Pillar ClassFine: trained on \dataname{CT-RATE} train labels ($\sim$39k volumes); encoder frozen.
$^\ast$\textbf{zero-shot:} ~\citet{hamamci2026generalist} present/absent prompt pairs ("$c$ is present."~/~"$c$ is not present.") embedded with each backbone's text encoder, scored against
cached $x_i$ via softmax over two cosines ---the
same recipe as cited \dataname{CT-CLIP zero-shot}, applied to \dataname{Pillar-0} (1152-d) or Merlin (512-d, Clinical-Longformer); no label training.
$^\ddagger$CCP-10: training-free channel probe + calibration set thresholds.
\textbf{Bold} = best supervised per column; \underline{\textbf{bold underline}} = best training-free per column.}
\label{tab:ctrate_bench}
\end{table}

\section{Experimental setup}
\label{sec:experiments}
\textbf{Data.} We evaluate on three public cohorts: 

\textbf{(1) \dataname{CT-RATE}}: 3D chest CTs with radiology reports and 
$C=18$ multi-label finding annotations. We use the official validation split 
($1564$ CTs) for test evaluation and a $600$-volume calibration set from 
\emph{train} for channel ranking, probe fitting, and threshold calibration.

\textbf{(2) \dataname{RadChest-CT}}~\cite{draelos_2020_6406114}: $3630$ 
external chest CTs with $84$ abnormality labels ($52$ location-based). Following 
the protocol of~\citet{hamamci2026generalist}, detectors trained on 
\dataname{CT-RATE} dev are evaluated without refitting, using 
\dataname{CT-RATE}'s 18-finding ontology.

\textbf{(3) \dataname{Merlin Abdominal CT}}~\cite{blankemeier2024merlin}: 
paired abdominal CTs with reports and $C=30$ multi-label finding annotations.

Each finding is coded $1$ (present), $0$ (absent), or $-1$ (not assessed); 
we treat $-1$ as missing during channel ranking and probe fitting.
% Official splits are \emph{train} ($xxxx$ volumes), \emph{valid} ($n{=}xxxx$),
% and \emph{test} ($n{=}xxxx$).
Because prevalence is highly imbalanced across findings, we fit a detector for any finding with at least five positive calibration volumes and abstain otherwise; on \dataname{CT-RATE} all $18$ findings qualify, while a few of the rarest \dataname{RadChest-CT} and \dataname{Merlin} native labels fall below this floor.

\noindent\textbf{3D VLM Backbones and $K$ selection} \label{sec:k_selection}
We run CCP on frozen precomputed vision-encoder embeddings from \dataname{Pillar-0} ($D=1152$) and \dataname{Merlin} ($D=512$). On \dataname{Pillar-0} channel ranking and probe directions are fit on a $600$-volume calibration set; $K$ is chosen on a disjoint $800$-volume train holdout by grid-searching $K\in\{1,2,5,10,20,50,100,200,D\}$ and taking the smallest $K$ within $0.005$ mean AUROC of full-$D$ ($K^*=10$). We use the same $K^*=10$ for \dataname{Merlin} without separate tuning.

\noindent\textbf{Classification metrics.}
F1, accuracy, and precision metrics follow \dataname{CT-CLIP} evaluation protocol (\citet{hamamci2026generalist}'s public evaluation code). Mean AUROC is threshold-free on raw scores. We report the numbers for \dataname{CT-CLIP} from the original paper since the public checkpoint is not available. 

\noindent\textbf{Report evaluation metrics.}
\label{sec:report_metrics}
For report generation, we measure (i) \textbf{Clinical efficacy:} micro/macro-F1 from a RadBERT-CT labeler on generated text vs.\ official \dataname{CT-RATE} volume labels (same GT as CCP), (ii) \textbf{Surface NLG:} BLEU-1 and ROUGE-L vs.\ reference reports, (iii) \textbf{Factual structure:} RadGraph-F1~\cite{delbrouck-etal-2024-radgraph}, and (iv) \textbf{Reference-aligned:} RadEval~\cite{xu-etal-2025-radeval} F1RadBERT-CT~\cite{wald2025comprehensive} on generated vs.\ reference report pairs.

\section{Results}
\subsection{Multi-label classification}
\label{sec:results_cls}
\label{sec:text_baseline}
Table~\ref{tab:ctrate_bench} summarizes results on \dataname{CT-RATE} and 
\dataname{RadChest-CT}. CCP uses frozen Pillar-0 embeddings. Baselines include: 
CT-CLIP zero-shot and ClassFine/VocabFine (cited from~\cite{hamamci2026generalist}); 
zero-shot prompts ("$c$ is present/absent") scored via text-encoder similarity; 
and Pillar-0 ClassFine, a supervised linear head.
CCP-10 is the best training-free method on both splits; supervised Pillar-0 
ClassFine leads overall. On Pillar-0, training-free CCP-10 ($0.798$) approaches 
supervised ClassFine ($0.833$) without label training, while zero-shot stays 
lower ($0.628$--$0.682$ AUROC). On unseen \dataname{RadChest-CT}, CCP-10 
remains best training-free (AUROC $0.686$, F1 $0.622$), retaining $\sim86\%$ 
in-domain AUROC vs. zero-shot. Supervised CT-CLIP VocabFine leads F1 ($0.677$); 
Pillar ClassFine reaches higher AUROC ($0.713$) but lower F1 ($0.609$).

\subsection{Training-free report generation}
We convert CCP-10 detections into reports on $1564$ test volumes, holding 
detections fixed and varying only the \emph{verbalizer}. Table~\ref{tab:report} 
reports clinical efficacy and NLG metrics.

We compare: \textbf{Structured template:} PRESENT/ABSENT lists formatted 
as FINDINGS/IMPRESSION. \textbf{Corpus-based template:} edits normal-study 
template $T_0$ with positive sentences $\{u_c\}$. \textbf{Frozen LLM:} 
off-the-shelf LLM (Qwen3-VL-8B-Instruct) generates reports from lists. 
\textbf{Retrieve-then-edit:} k-NN retrieves closest training report, then 
edits to match $d_i$. \textbf{k-NN constrained:} k-NN retrieval limited to 
normal reports, then edit to match $d_i$. \textbf{CT-CHAT:} trained chest CT 
VLM~\cite{hamamci2026generalist}.

\begin{table}[t]
\centering
\small
\setlength{\tabcolsep}{1pt}
\renewcommand{\arraystretch}{1}
\begin{tabular}{l|c|c|c|c|c|c}
\toprule
Method & Tr. & $\mu$F1 & M-F1 & B1 & R-L & Lt. \\
\midrule
Structured template        & no  & .441 & .408 & .015 & .064 & \best{.24} \\
Frozen LLM   & no  & .315 & .317 & .057 & .081 & 1.15 \\
\best{Courpus-based Template.}& no  & \best{.549} & \best{.492} &
\best{.483} & \best{.379} & \best{.24} \\
Retrieval-edit         & no  & .540 & .484 & .439 & .314 & .24 \\
$k$-NN retrieval   & no  & .544 & .486 & .414 & .297 & .24 \\
CT-CHAT$^\dag$     & yes & .184 & -- & .373 & .326 & 5.5 \\
\midrule
Radiologist        & -- & .962 & .944 & -- & -- & -- \\
\bottomrule
\end{tabular}
\caption{Report generation on \dataname{CT-RATE} test ($n{=}1564$).
Clinical Micor-F1 ($\mu$F1), Macro-F1 (M-F1), evaluated by F1RadBERT-CT vs.\ official labels.
NLG: BLEU-1 (B1) / ROUGE-L (R-L) vs.\ radiologist FINDINGS. Lateny (Lt.) as wall-clock time to produce one report end-to-end based on second per volume (s/v).
CCP-10 corpus-derived template verbalizer is based on \dataname{Pillar-0}.
$^\dag$\dataname{CT-CHAT}: extracted from ~\citet{hamamci2026generalist}.}
\label{tab:report}
\end{table}

The CCP-10 corpus-based template is the strongest \emph{training-free} verbalizer 
on all three axes: RadBERT-CT F1 $0.549$ micro / $0.492$ macro (vs.\ cited 
CT-CHAT $0.184$), BLEU-1 $0.483$ / ROUGE-L $0.379$ (vs.\ CT-CHAT $0.373$ / 
$0.326$), and latency $0.24$\,s/vol—$\sim$23$\times$ faster than CT-CHAT 
($5.5$\,s/vol). Retrieve-then-edit matches clinical F1 ($0.540$) but drops 
NLG (BLEU $0.439$, ROUGE-L $0.314$) due to non-corpus phrasing. Structured 
template and frozen LLM sacrifice prose style (BLEU $\leq 0.064$ and 
$\sim$0.01 respectively). Pairwise bootstrap confirms retrieve-then-edit and 
k-NN lag corpus F1 ($p<0.001$), with corpus F1 in tight 95\% CI 
(Table~\ref{tab:bootstrap_stats}).

\paragraph{Statistical analysis.}
\label{sec:bootstrap}
We assess uncertainty via paired volume bootstrap ($n=1564$ test volumes, 
$B=2000$ resamples; calibration/dev excluded). Point estimates report 95\% 
bootstrap confidence intervals. Pairwise comparisons use two-sided bootstrap 
$p$-values with shared resampling indices. Classification uses mean 
macro-AUROC (threshold-free); report generation uses RadBERT-CT micro-F1 
(Sec.~\ref{sec:report_metrics}). Table~\ref{tab:bootstrap_stats} summarizes 
primary comparisons.
% Cited CT-CHAT/CT-CLIP baselines are excluded from bootstrapping (different checkpoints/splits).

\begin{table}[t]
\centering
\small
\begin{tabular}{@{}lccc@{}}
\toprule
Comparison & $\Delta$ & 95\% CI & $p$ \\
\midrule
\multicolumn{4}{@{}l}{\textit{Report generation (Clin-F1 vs.\ corpus template)}} \\
\multicolumn{4}{@{}l@{}}{Corpus template: $0.549$ [0.537, 0.560] (absolute CI).} \\
Structured template & -0.078 & [-0.086, -0.071] & $p<0.001$ \\
Retrieve-then-edit & -0.009 & [-0.013, -0.004] & $p<0.001$ \\
$k$-NN retrieval & -0.005 & [-0.007, -0.002] & $p<0.001$ \\
Frozen LLM (s2t) & -0.233 & [-0.245, -0.221] & $p<0.001$ \\
\bottomrule
\end{tabular}
\caption{Paired bootstrap tests on \dataname{CT-RATE} test ($n=1564$, $B=2000$). Clin-F1: F1RadBERT-CT micro-F1; classification: mean macro-AUROC. Reference row for verbalizers: corpus template (CCP-10).}
\label{tab:bootstrap_stats}
\end{table}

\begin{table}[t]
\centering
\footnotesize
\setlength{\tabcolsep}{2pt}

\noindent\textbf{Classification transfer.}
\vspace{0.2em}

\begin{tabular}{@{}llcc@{}}
\toprule
Sce. & \hspace{10pt}Calibration $\rightarrow$ Test & AUROC &  $\mu$-F1 \\
\midrule
\multicolumn{4}{@{}l}{\textit{cross-institution (\dataname{Pillar-0} chest)}} \\
A & \dataname{RadChest-CT} valid $\rightarrow$ \dataname{CT-RATE} test & 0.726 & 0.462 \\
A & \dataname{CT-RATE} dev $\rightarrow$ \dataname{RadChest-CT} all & 0.686 & 0.622 \\
\addlinespace
\multicolumn{4}{@{}l}{\textit{cross-anatomy/backbone~\dataname{Merlin}}} \\
B &  abd. CT\ valid $\rightarrow$  abd. CT\ test ($30$ findings) & \besttf{0.839} & \besttf{0.832} \\
--- & zero-shot$^{\dagger}$ & --- & 0.647 \\
--- & supervised$^{\dagger}$ & --- & 0.641 \\
\addlinespace
\multicolumn{4}{@{}l}{\textit{anatomy mismatch, \dataname{Merlin} encoder/Chest labels}} \\
C & \dataname{CT-RATE} train $\rightarrow$ \dataname{CT-RATE} test & 0.733 & 0.733 \\
C & \dataname{CT-RATE} train $\rightarrow$ \dataname{RadChest-CT}  & 0.605 & \tna \\
\bottomrule
\end{tabular}

\vspace{0.5em}
\noindent\textbf{Report generation transfer.}
\footnotesize
\setlength{\tabcolsep}{1.5pt}
\begin{tabularx}{\linewidth}{@{}c >{\raggedright\arraybackslash}X ccccc@{}}
\toprule
Sce. & \hspace{10pt}Calibration $\rightarrow$ Test 
& F1 & RG-F1 & B-1 & R-L & BERT \\
\midrule

\multicolumn{7}{@{}l}{\textit{cross-institution} (\dataname{Pillar-0} chest)} \\[0.2ex]
A & \dataname{RadChest-CT} valid $\rightarrow$ \dataname{CT-RATE} test
& 0.448 & 0.284 & 0.471 & 0.352 & \tna \\

\multicolumn{7}{@{}l}{\textit{\dataname{Merlin} abdomen}} \\[0.2ex]
B & abd.\ valid $\rightarrow$ abd. test
& \tna & 0.262 & \besttf{0.214} & 0.226 & \besttf{0.677} \\

--- & Merlin+RadLlama$^\dagger$
& \tna & \besttf{0.293} & 0.102 & \tna & 0.588 \\

\bottomrule
\end{tabularx}
\caption{CCP-10 transfer summary. $^\dag$~extracted from~\citet{hamamci2026generalist,blankemeier2024merlin}}
\label{tab:transfer_overview}
\end{table}
\subsection{Transfer across datasets, backbones, and anatomy}
\label{sec:transfer}
Since our concept channels probe~(CCP) is defined entirely on frozen embeddings, the same recipe ($K=10$) can be applied under three distinct transfer scenarios (Table~\ref{tab:transfer_overview}):
\textbf{(A)~Cross-institution}: we use the same VLM backbone and chest anatomy, but different hospital/label ontology
(\dataname{Pillar-0} on \dataname{CT-RATE} vs.\ \dataname{RadChest-CT};
\textbf{(B)~Cross-anatomy/backbone}: we apply the same CCP recipe to another 3D VLM backbone as \dataname{Merlin} on \dataname{Abdominal CT} dataset~\cite{blankemeier2024merlin};
\textbf{(C)~Anatomy mismatch}: we investigate whether the \emph{chest} finding labels probed on an \emph{abdominal}-pretrained
\dataname{Merlin} encoder can be applied to \dataname{CT-RATE} and \dataname{RadChest-CT}.
For each scenario we rank channels, fit probe directions, and calibrate thresholds entirely on the \emph{calibration} split of the source dataset, followed by evaluation on the \emph{test} target data split. Report generation metrics utilize a fixed, corpus-derived template verbalizer, varying only the underlying detection source.

\textbf{Scenario~(A): \dataname{Pillar-0} cross-institution.}
\label{sec:radchest}
We ask whether CCP probes \emph{fit at one hospital} transfer to the other while keeping the frozen 3D VLM backbone~(\dataname{Pillar-0}) fixed.
% On the chest findings mapped between \dataname{CT-RATE} and \dataname{RadChest-CT}, we rank channels, fit probe directions, and calibrate thresholds on the \textbf{source} calibration split only, then evaluate on the \textbf{target} institution test set without refitting weights or retuning on target data.
Probes are \emph{fit} in each hospital's native label space ($18$ on \dataname{CT-RATE} dev; $84$ binary on \dataname{RadChest-CT}), but we \emph{evaluate} on the target only through the mapped \dataname{CT-RATE} ontology similar to \citet{hamamci2026generalist}.
In \textbf{classification}, as shown in Table~\ref{tab:transfer_overview} (scenario A), the bidirectional transfer is asymmetric but strong \dataname{RadChest-CT}$\rightarrow$\dataname{CT-RATE} test reaches AUROC $0.726$ and F1 $0.462$;
\dataname{CT-RATE}$\rightarrow$\dataname{RadChest-CT} all volumes reaches $0.686$ / $0.622$ (also reported in  Table~\ref{tab:ctrate_bench}). In 
\textbf{report generation}, Table \ref{tab:transfer_overview} (scenario A), we keep the \dataname{CT-RATE} corpus template and test split fixed and swap only the detection source to probes calibrated on \dataname{RadChest-CT}, which yields F1 $0.448$, BLEU $0.471$, and ROUGE-L $0.352$ vs. radiologist FINDINGS. It can be seen that without 
\dataname{CT-RATE} detector labels, cross-institution transfer retains $\sim82\%$ of F1 in-domain report generation (see \ref{tab:report}) and still exceeds CT-CHAT performance.

\textbf{Scenario~(B): \dataname{Merlin} cross-anatomy/backbone}
\label{sec:radchest}
We apply CCP-10 to abdominal CT classification by ranking sparse channels in frozen Merlin contrastive embeddings, and calibrating per-finding probe directions and thresholds on the valid split with $30$ findings, and evaluating the frozen channels on the test set. As shown in Table~\ref{tab:transfer_overview} (scenario A), in \textbf{classification} task, CCP-10 reaches AUROC $0.839$ and F1 $0.832$, exceeding \dataname{Merlin} zero-shot and supervised~\cite{blankemeier2024merlin} $0.647$ and  $0.641$ F1, respectively. In \textbf{report generation (Table~\ref{tab:transfer_overview}, scenario B):} with the same CCP-10 detections, we mine a corpus template from Merlin \emph{train} FINDINGS, apply the same deterministic edit rule. Our proposed method yields RadGraph-F1 (RG-F1) $0.262$, BLEU-1 (B1) $0.214$, ROUGE-L (R-L) $0.226$, and BERTScore F1 (BERT) $0.677$.
Compared to Merlin+RadLlama~\cite{blankemeier2024merlin}, our training-free method improves BLEU-1 and BERTScore ($0.214$ and
$0.677$ vs.\ $0.102$ and $0.588$) but yields lower RadGraph-F1 ($0.262$ vs.\ $0.293$).

\textbf{Regime~(C): Abdominal encoder~\dataname{Merlin} on chest Labels. anatomy mismatch }
\label{sec:regime_c}
We apply the same CCP-$10$ recipe to \dataname{CT-RATE} chest findings but swap the frozen encoder
to abdominal-pretrained \dataname{Merlin}---an intentional anatomy mismatch that complements
\citet{blankemeier2024merlin}'s external benchmarks, where matched-anatomy zero-shot F1 drops from $0.741$ to $0.647$ under hospital \emph{distribution shift} yet frozen \dataname{Merlin} still generalizes to external
chest CT in their linear-probe evaluation~\cite{blankemeier2024merlin}.
Probes rank channels, fit directions, and calibrate thresholds on \dataname{CT-RATE} dev, then evaluate frozen readouts on official \dataname{CT-RATE} test ($n{=}1564$) and on all \dataname{RadChest-CT}
volumes ($n{=}3630$; mapped ontology). In \textbf{classification (Table~\ref{tab:transfer_overview}, regime C)}, we can see on \dataname{CT-RATE} test, CCP-10 reaches AUROC $0.733$ and F1 $0.733$; near original \dataname{CT-CLIP} zero-shot ($0.731$ / $0.707$; but below chest-native \dataname{Pillar-0} CCP-10 ($0.798$ / $0.790$). Cross-institution evaluation on \dataname{RadChest-CT} all volumes yields AUROC $0.605$, well below regime~(A) \dataname{Pillar-0} transfer ($0.686$ / $0.622$). 
Regime~(C) therefore complements \dataname{Merlin}'s external chest benchmarks~\cite{blankemeier2024merlin}, and we observe that the CCP procedure transfers across backbones, but encoder pretraining and probe labels must align, as shown in Scenarios A and B. 

\section{Analysis and discussion}
\label{sec:discussion}
CCP-10 succeeds on frozen chest and abdominal encoders. Here we analyze 
\emph{where} that signal lives in the embedding. We ask: (i) Are findings encoded 
by sparse channels? (ii) Are these channels causally necessary? (iii) Does 
this hold across anatomies?

\begin{figure}[h!]
\centering
\includegraphics[width=\linewidth]{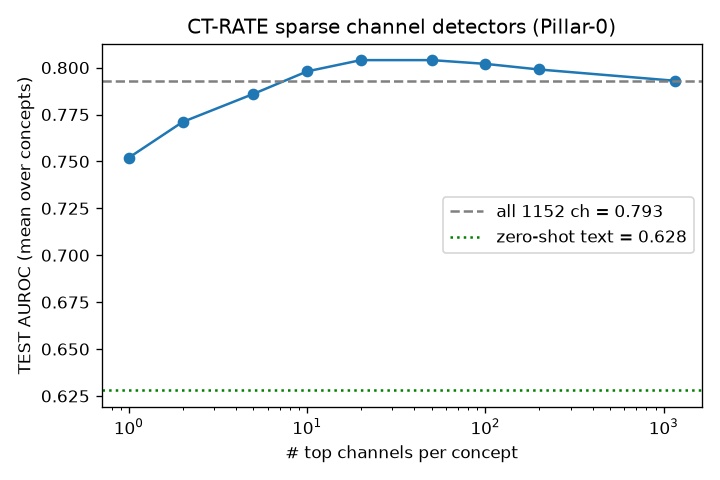}
\caption{Sparsity recovery: test classification AUROC vs.\ number of top channels per finding.
A handful of channels recover most of the full-feature performance on \dataname{Pillar-0}.}
\label{fig:sparsity}
\includegraphics[width=\linewidth]{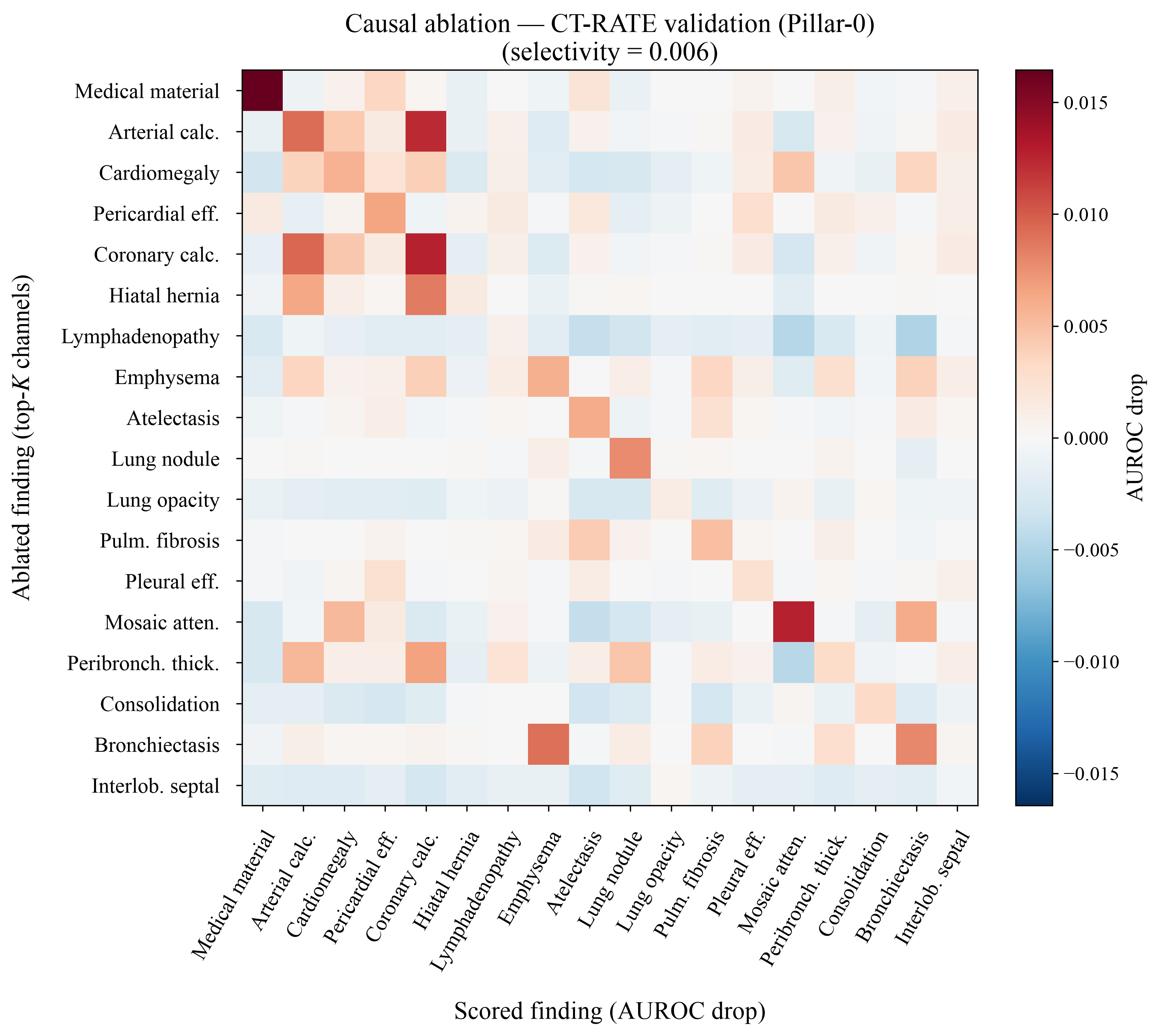}
\captionof{figure}{Causal ablation on \dataname{CT-RATE} validation with frozen \dataname{Pillar-0}.
Row~$c$: AUROC changes after zeroing source finding~$c$'s dev-ranked top-$K{=}10$ circuit ($\mathcal{S}^{\mathrm{abl}}_c=\mathcal{S}^K_c$); column~$c'$: scored finding. Cell $(c,c')$ shows $\Delta_{c\to c'}$ from Eq.~\ref{eq:abl_formal}. A strong diagonal indicates selective necessity; off-diagonal blocks indicate shared circuits among related findings.}
\label{fig:ablation}
\end{figure}

\textbf{Each finding lives in $\sim$10 channels.}
We rank channels by per-concept AUROC on the calibration set, apply CCP 
with varying top-$K$, and evaluate on $1564$ test volumes at $K=10$. 
%Fig.~\ref{fig:sparsity} plots validation AUROC vs. $K$.
% Table~\ref{tab:sparse} and Fig.~\ref{fig:sparsity} summarize the $K$-sweep on test (mean AUROC
% over 18 findings using top-$K$ channels, with $K$
% varying as shown).
% \best{Top-${10}$ matches the full 1152-d readout} ($0.798$ vs.\ $0.793$) while using only
% $10$ coordinates per finding ($\sim$100$\times$ fewer than $D{=}1152$), and \best{beats zero-shot} on \dataname{Pillar-0} ($0.682$).
% Performance rises sharply from top-1 ($0.752$) to top-$10$ ($0.798$) and saturates by top-50
% ($0.804$), It shows that the most discriminative signal for each finding concentrates in a small channel subset
% rather than requiring the full embedding.
% Gains beyond the top-50 are marginal.
% \begin{table}[h]
% \centering
% \begin{tabular}{lccccc}
% \toprule
% detector & top-1 & top-10 & top-50 & all & zero-shot \\
% \midrule
% \dataname{Pillar-0} & 0.752 & 0.798 & 0.804 & 0.793 & 0.682 \\
% % \dataname{Merlin}   & 0.695 & 0.729 & 0.728 & 0.733 & 0.571 \\
% \bottomrule
% \end{tabular}
% \caption{Sparse channel detectors on \dataname{CT-RATE} (test AUROC, mean over 18 findings).
% Ten channels per finding match the full feature and beat zero-shot text on \dataname{Pillar-0}.}
% \label{tab:sparse}
% \end{table}
Fig.~\ref{fig:sparsity} shows mean test AUROC across 18 \dataname{CT-RATE} findings 
vs. top-$K$ channels on \dataname{Pillar-0}. Performance rises sharply from top-1 
($0.752$) to top-10 ($0.798$), saturates by top-50 ($0.804$), and reaches 
$0.793$ with all $D=1152$ coordinates. Top-10 matches full-dimension AUROC 
($0.798$ vs. $0.793$) using only $10$ coordinates per finding ($\sim$100× 
fewer), and beats zero-shot text prompting ($0.682$). Top-10 can exceed 
full-embedding AUROC on individual findings (e.g., lung nodule $0.734$ vs. 
$0.683$, atelectasis $0.746$ vs. $0.697$) by dropping noisy dimensions. 
Table~\ref{tab:per_finding_sparsity} lists each finding with its channel 
set $\mathcal{S}^{10}_c$ and test AUROC.
%Fig.~\ref{fig:sparsity} plots the $K$-sweep of mean test AUROC (over the $18$ \dataname{CT-RATE} findings) using the top-$K$ channels per finding on \dataname{Pillar-0}. Performance rises sharply from top-$1$ ($0.752$) to top-$10$ ($0.798$), then saturates by top-$50$ ($0.804$) with only marginal gains beyond, and stays at $0.793$ when using all $D{=}1152$ coordinates. \best{Top-$10$ matches the full $1152$-d readout} ($0.798$ vs.\ $0.793$) while using only $10$ coordinates per finding ($\sim$100$\times$ fewer than $D$), and \best{beats zero-shot} text prompting on \dataname{Pillar-0} ($0.682$). This shows that the most discriminative signal for each finding concentrates in a small channel subset rather than requiring the full embedding. Table~\ref{tab:per_finding_sparsity} lists each \dataname{CT-RATE} label with its identified channel set $\mathcal{S}^{10}_c$ and AUROC on test set.
%On \dataname{Pillar-0}, mean AUROC is $0.798$ with top-$K^*$ vs.\ $0.793$ with all $1152$ coordinates, the
%sparse circuit matches the full feature on average and can \emph{exceed} it for individual findings
%(e.g.\ lung nodule $0.734$ vs.\ $0.683$, atelectasis $0.746$ vs.\ $0.697$) because the CCP-ranked coordinates drop noisy dimensions.
Gaps are modest in both directions ($|\Delta\mathrm{AUROC}|\le 0.05$ for all 18 labels). Hiatal hernia shows the largest gap (top-10: $0.027$ below full). 
Fluid findings stay near ceiling (pleural effusion $0.928/0.944$, 
consolidation $0.905/0.886$). 

\textbf{Causal ablation: Do sparse channels matter?}
We show that a few channels ($K=10$) correlate strongly with each finding; however, correlation alone does not establish that the selected coordinates are necessary for discrimination. On \dataname{CT-RATE} test set (\dataname{Pillar-0}, $n{=}1564$), we zero $\mathcal{S}^{\mathrm{abl}}_c=\mathcal{S}^K_c$ (Eqs.~\ref{eq:topk_formal},~\ref{eq:abl_formal}) for each source finding~$c$ and record $\Delta_{c\to c'}$ for every scored finding~$c'$. 
Figure~\ref{fig:ablation} displays the resulting $18{\times}18$ matrix with source~$c$ on rows and scored~$c'$ on columns (Eq.~\ref{eq:abl_formal}). We ablate the same dev-ranked top-$K$ channels that CCP uses for each finding
($\mathcal{S}^{\mathrm{abl}}_c=\mathcal{S}^K_c$). If those coordinates encoded only generic shared signal, zeroing finding~$c$'s circuit would lower every detector by about the same amount; instead, the mean on-target AUROC drop is $0.0060$ versus $0.00003$ off-target ($\sim$20$\times$), so each finding's channels are \emph{selectively necessary} for its own score rather than broadly shared across all labels. Some off-diagonal entries remain elevated among clinically related findings (e.g.,\ arterial and coronary calcification; emphysema with atelectasis and lung opacity), consistent with partial circuit overlap among related pathologies.

%There is a large shortfall in hiatal hernia, top-$10$ $0.027$ below full). Strong fluid and airspace findings stay near the ceiling (pleural effusion $0.928/0.944$, consolidation $0.905/0.886$).
% \begin{figure}[h]
% \centering
% \includegraphics[width=\linewidth]{Figures/pilar-k.png}
% \caption{Sparsity recovery: test classification AUROC vs.\ number of top channels per finding.
% A handful of channels recover most of the full-feature performance on \dataname{Pillar-0}}
% \label{fig:sparsity}
% \end{figure}

\begin{table}[t]
\centering
\fontsize{6.5pt}{7.5pt}\selectfont % Custom compact font size and line spacing
\setlength{\tabcolsep}{3pt}        % Tighten horizontal spacing
\renewcommand{\arraystretch}{0.85} % Compress vertical row padding
\begin{tabularx}{\columnwidth}{@{}l X cc@{}}
\toprule
Finding & $\mathcal{S}^{10}_c$ & top-$10$ & all $D$ \\
\midrule
medical material & 525, 923, 941, 904, 589, 1110, 848, 429, 870, 457 & 0.889 & 0.849 \\
arterial wall calcification & 792, 897, 1145, 564, 588, 607, 490, 981, 889, 97 & 0.864 & 0.863 \\
cardiomegaly & 897, 1103, 1145, 813, 423, 607, 792, 1045, 761, 429 & 0.878 & 0.863 \\
pericardial effusion & 552, 627, 761, 570, 1087, 923, 560, 872, 609, 446 & 0.813 & 0.767 \\
coronary artery wall calcification & 792, 1145, 588, 564, 490, 607, 897, 889, 407, 981 & 0.879 & 0.866 \\
hiatal hernia & 1078, 875, 607, 981, 1126, 457, 521, 588, 490, 407 & 0.652 & 0.679 \\
lymphadenopathy & 897, 1137, 1045, 1072, 813, 549, 985, 604, 634, 901 & 0.713 & 0.721 \\
emphysema & 771, 1100, 1009, 385, 792, 1107, 716, 429, 1069, 545 & 0.760 & 0.770 \\
atelectasis & 751, 584, 586, 623, 483, 613, 851, 968, 1039, 967 & 0.746 & 0.697 \\
lung nodule & 643, 508, 661, 663, 557, 386, 462, 528, 442, 543 & 0.734 & 0.683 \\
lung opacity & 404, 595, 721, 719, 517, 648, 743, 707, 594, 602 & 0.856 & 0.858 \\
pulmonary fibrotic sequela & 559, 586, 578, 613, 751, 454, 680, 1066, 727, 1064 & 0.682 & 0.692 \\
pleural effusion & 570, 609, 446, 395, 552, 761, 498, 659, 560, 444 & 0.928 & 0.944 \\
mosaic attenuation pattern & 897, 1103, 1007, 879, 858, 813, 784, 894, 982, 1045 & 0.792 & 0.786 \\
peribronchial thickening & 885, 564, 239, 1069, 429, 771, 457, 484, 904, 792 & 0.717 & 0.744 \\
consolidation & 689, 426, 648, 413, 551, 549, 600, 901, 730, 418 & 0.905 & 0.886 \\
bronchiectasis & 781, 716, 454, 1107, 1000, 961, 545, 397, 1100, 577 & 0.727 & 0.753 \\
interlobular septal thickening & 743, 1097, 1072, 428, 740, 482, 1013, 927, 719, 721 & 0.823 & 0.861 \\
\bottomrule
\end{tabularx}
\caption{Per-finding CCP-$K^*$ circuits on \dataname{Pillar-0} (\dataname{CT-RATE} validation, $n{=}1564$, $D{=}1152$). For each finding, $\mathcal{S}^{K^*}_c$ lists dev-ranked channel indices (Eq.~\ref{eq:topk_formal}); top-$K^*$ and all-$D$ columns are threshold-free test ROC-AUC.}
\label{tab:per_finding_sparsity}
\end{table}

% \paragraph{Sparse, causally selective channels.}
% On frozen \dataname{Pillar-0}, each \dataname{CT-RATE} finding is carried by a small top-$K$ channel subset, typically $K=10$, that matches the full $1152$-d embedding for classification~(Fig.~\ref{fig:sparsity}).
% This effect cannot be explained as mere compression: when we zero out a circuit associated with a particular finding, the probe AUROC for that finding collapses, while AUROCs for unrelated labels remain nearly unchanged—differing by up to two orders of magnitude (Fig.~\ref{fig:ablation}).
% This is not simply a compression trick, since when we zero out a circuit that is associated with a particular finding collapses its own probe AUROC while AUROCs for unrelated findings remain nearly unchanged (the gap is up to two orders of magnitude)(Fig.~\ref{fig:ablation}). On the other hand, related findings share partial overlap on their circuits (calcification pairs, airspace disease) in ways that mirror clinical taxonomy.
\begin{figure}[t]
\centering
\includegraphics[width=1\linewidth]{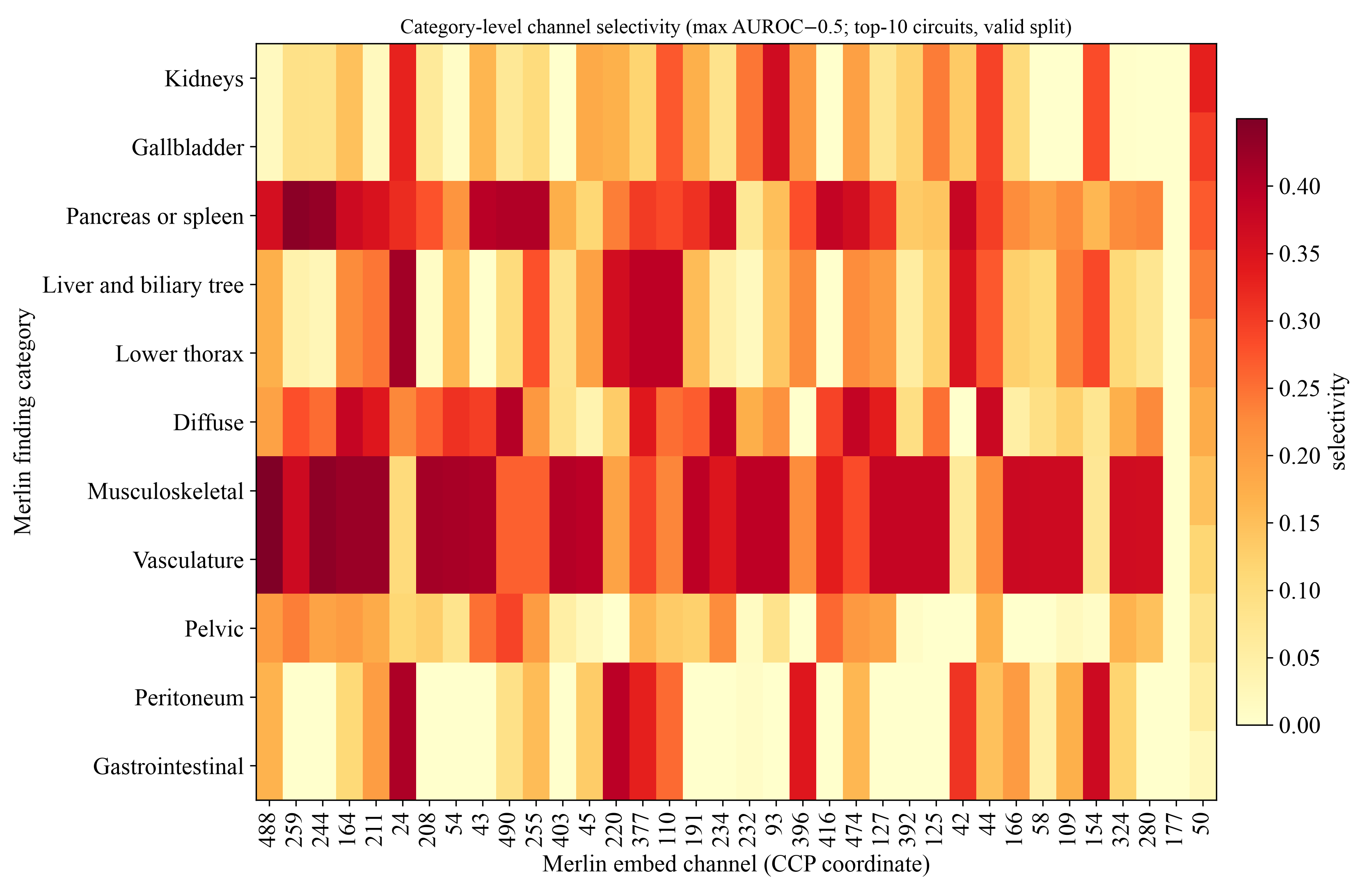}
\caption{\textbf{Merlin abdominal CCP circuits by finding category (Figure 2c grouping).}
Each row is one of Merlin's $11$ clinical categories ($30$ findings total).
Columns show embed coordinates with highest category-level selectivity (max AUROC$-0.5$ across
findings in that category; valid split).
Color intensity = selectivity. }
\label{fig:merlin_abd_sections}
\end{figure}

\textbf{Merlin circuits by clinical category.}
We aggregate the top-$10$ CCP circuits into $11$ finding categories, following~\citet{blankemeier2024merlin}. Fig.~\ref{fig:merlin_abd_sections}).
Category-level selectivity (max over findings of $\tilde{A}_{c,j}=\mathrm{AUROC}_{c,j}-0.5$ per coordinate) localizes sparse channels to clinically meaningful anatomy~(e.g., lower-thorax calcification and cardiac findings share coordinates, whereas pleural effusion uses a separate circuit; peritoneal ascites and free air reach mean top-$10$ AUROC $0.895$. Vasculature findings, by contrast, show heterogeneous top-$10$ sets, consistent with diverse vascular appearances. Cross-category Jaccard overlap is near zero for unrelated groups (liver vs. lower thorax) but rises where pretraining anatomy overlaps (pancreas/spleen vs.\ lower thorax, $0.43$), mirroring
the partial circuit sharing among related chest findings in Fig.~\ref{fig:ablation}.
% We observe the same pattern on \dataname{Merlin} abdominal CT, where category-level circuits localize to liver, kidney, peritoneum, and lower-thorax groupings (Fig.~\ref{fig:merlin_abd_sections}).
Together, these results suggest contrastive 3D pretraining concentrates finding information in sparse coordinates rather than spreading it uniformly. CCP localizes, thresholds, and ablates these coordinates without training.%that contrastive 3D pretraining does not spread finding information uniformly across the vision embedding; instead, it concentrates concept signal in sparse coordinates that CCP can name, threshold, and intervene on training-free.

\textbf{Interpretability vs.\ supervised finetuning.}
Supervised finetuning (CT-CLIP ClassFine/VocabFine, Pillar ClassFine; 
Table~\ref{tab:ctrate_bench}) optimizes predictions by updating decision 
boundaries. CCP localizes \emph{where} information lives by ranking channels and 
fitting a closed-form probe on frozen weights.
On CT-RATE, CCP-10 exceeds reported CT-CLIP zero-shot and remains the best 
training-free probe, uniquely enabling sparsity and ablation analysis. On 
RadChest-CT, CCP-10 leads AUROC but lags VocabFine on weighted F1. 
Finetuning optimizes scores; CCP audits deployed encoders and provides 
channel-level explanations alongside predictions.

\textbf{Detection-grounded reports.}
We separate detection (CCP) from verbalization, avoiding end-to-end VLM 
failure modes. CT-CHAT achieves BLEU-1 $0.373$ but F1 only $0.184$, drifting 
toward normal wording and missing abnormalities. CCP-10 + corpus template wins 
on both metrics (F1 $0.549$, BLEU-1 $0.483$) at $\sim$23$\times$ lower latency.
Two checks support this design. Unrelated reports score BLEU-1 $0.385$ against 
each other, so BLEU reflects boilerplate, not correctness. Corpus templates 
preserve institutional negations better than frozen LLMs. Retrieve-then-edit 
lags corpus template on F1, showing that with trustworthy detections, the 
bottleneck is faithful reconciliation to $d_i$, not generation.
On abdominal CT, CCP-10 + Merlin template improve NLG scores over 
Merlin+RadLlama without fine-tuning. Trained RadLlama achieves higher 
RadGraph-F1 through richer entity encoding.

\paragraph{Transfer across institutions, backbones, and anatomy.}
The three regimes in Table~\ref{tab:transfer_overview} form a single 
generalization story:
Regime~(A): Sparse chest circuits from one hospital retain signal at another 
(AUROC $0.686$ / $0.726$; report F1 $\sim$82\% of in-domain). Institutional 
distribution shift does not erase channel structure.
Regime~(B): The CCP recipe is backbone-portable. On abdominal \dataname{Merlin}, 
CCP exceeds Merlin zero-shot and supervised F1 ($0.832$ vs.\ $0.647$ / $0.641$).
Regime~(C): Chest probes on abdominal \dataname{Merlin} remain in-domain 
($0.733$) but lag chest-native Pillar-0 and degrade under hospital shift 
($0.605$ vs.\ $0.686$). Encoder and probe anatomy must align; the ranking 
procedure transfers across backbones, not across mismatched anatomies.
% In practice we read out chest \dataname{CT-RATE} and \dataname{RadChest-CT} through \dataname{Pillar-0}
% and abdominal volumes through \dataname{Merlin} using regime~(C) only as a stress test, not a recommended setting.

This work presents one consistent picture: Frozen 3D CT encoders store findings 
in sparse causally selective channel subsets. CCP localizes and names these 
coordinates training-free, whereas finetuning improves \emph{what} is predicted but 
not \emph{where} it is encoded. CCP-10 achieves competitive performance while 
providing interpretability, grounding language in reliable detections rather 
than free-form generation.
The probe-and-readout recipe transfers across hospitals and backbones when 
anatomy matches, and breaks down when it does not (regime~C). For deployed 
foundation models, AUROC alone is an incomplete account of trust. Users need 
assurance that findings rest on stable internal features. CCP audits deployed 
models by probing, naming, ablating, and verbalizing.
A limitation: rare findings have higher-variance channel estimates (few positive 
calibration volumes). We mitigate by reporting threshold-free AUROC, using 
prevalence-independent ROC operating points, and excluding low-count findings. 
Whether channel-level explanations remain stable under finer pathology labels, 
longitudinal follow-up, and multimodal inputs remains future work.

\section{Conclusion}
In this work we show how frozen medical VLMs encode radiological findings in sparse, causally selective channels. CCP localizes, ablates, and converts these into predictions and reports on chest (\dataname{Pillar-0}) and abdominal (\dataname{Merlin}) settings. The probe-and-template pipeline is competitive with finetuned VLMs on classification and report metrics while operating at a fraction of the latency.
CCP can serve as a fast perception module in multi-agent workflows, supplying 
structured detections and channel evidence to routing agents, reducing latency 
in multi-step pipelines.
We view CCP as a practical route to auditing \emph{what} and \emph{where} encoders 
represent findings, enabling lightweight, repeated inference without full VLM cost.
Future work should test whether channel circuits remain stable under finer 
pathology labels, longitudinal follow-up, and multimodal inputs. We will also 
evaluate CCP inside multi-agent benchmarks with latency and interpretability 
as first-class metrics.

\bibliography{aaai2027}

\end{document}